\documentclass[lettersize,journal]{IEEEtran}
\usepackage{amsmath,amsfonts}
\usepackage{algorithmic}
\usepackage{algorithm}
\usepackage{array}
\usepackage[caption=false,font=normalsize,labelfont=sf,textfont=sf]{subfig}
\usepackage{textcomp}
\usepackage{stfloats}
\usepackage{url}
\usepackage{verbatim}
\usepackage{graphicx}
\usepackage{cite}
\usepackage{booktabs}
\usepackage{longtable}
\usepackage{multirow}
\usepackage{makecell}
\usepackage{color}
\IEEEpubidadjcol 

\begin{document}
\title{STATrack: A Target‑Aware Fully Spiking Neural Network for Efficient UAV Tracking}

\author{Pengzhi Zhong, Jiwei Mo, Dan Zeng, Feixiang He, and Shuiwang Li$^*$\thanks{*Corresponding author.}

\thanks{Pengzhi Zhong, Jiwei Mo, Shuiwang Li are with the Guilin University of Technology, Guilin 541006,
China. (e-mail: zhongpengzhi@glut.edu.cn; mjwmm@glut.edu.cn; 
lishuiwang0721@glut.edu.cn).}
\thanks{Dan Zeng is with School of Artificial Intelligence, Sun Yat-sen University, Zhuhai 510275, China (e-mail: zengd8@mail.sysu.edu.cn).}
\thanks{Feixiang He is with School of Electronic Information, Central South University, Changsha, Hunan, 410083, China (e-mail: feixiang.he@csu.edu.cn).}
}

\markboth{Journal of \LaTeX\ Class Files,~Vol.~XX, No.~X, XX~202X}%
{Shell \MakeLowercase{\textit{et al.}}: A Sample Article Using IEEEtran.cls for IEEE Journals}

\IEEEpubidadjcol
\maketitle

\begin{abstract}
Spiking Neural Networks (SNNs), characterized by their event-driven computation and low power consumption, have shown great potential for energy-efficient visual tracking on unmanned aerial vehicles (UAVs). However, existing SNN-based trackers often rely on costly event cameras, which limits their deployment on standard RGB-camera UAV platforms. To address this limitation, we propose STATrack, a fully spiking neural network framework for UAV visual tracking using only RGB inputs. To the best of our knowledge, this is the first study to explore fully spiking neural networks for RGB-based UAV visual tracking. Considering the challenge of preserving fine-grained target information in deep spiking representations under a limited number of time steps, we introduce Adaptive Mutual Information Maximization (AMIM). AMIM maximizes the statistical dependency between target-centered templates and their deep spiking representations, encouraging the spiking backbone to learn representations that better distinguish the target from surrounding backgrounds. We further develop a sample-difficulty-aware dynamic weighting strategy that adaptively adjusts the mutual-information constraint according to the relative localization difficulty of each training batch. Extensive experiments on four widely used UAV tracking benchmarks demonstrate that STATrack achieves state-of-the-art tracking performance with low theoretical energy consumption, highlighting its potential for energy-constrained UAV applications. Code is released at: \url{https://anonymous.4open.science/r/STATrack}. 
\end{abstract}


\section{Introduction}

Unmanned aerial vehicle (UAV) visual tracking is a critical task in computer vision, with broad applications in intelligent surveillance, emergency rescue, smart agriculture, and aerial photography \cite{ARCF,HiFT,AVTrack,AbaViTrack}. Compared to generic object tracking, UAV tracking faces more severe challenges, including extreme viewpoint variations, significant motion blur, and complex background clutter \cite{AutoTrack,TATrack,ORTrackDeiT}. Moreover, UAV platforms are strictly constrained by battery capacity and onboard computational resources, which impose stringent requirements on both efficiency and energy consumption. Although recent tracking approaches based on deep convolutional neural networks (CNNs) and vision Transformers (ViTs) have achieved remarkable improvements in accuracy, their high computational complexity and energy consumption still hinder long-term real-time deployment on embedded UAV systems \cite{AbaViTrack,ORTrackDeiT}. Therefore, reducing computational energy consumption while maintaining high tracking accuracy remains a key challenge in UAV visual tracking.
\begin{figure}
	\centering
    \includegraphics[width=0.98\columnwidth]{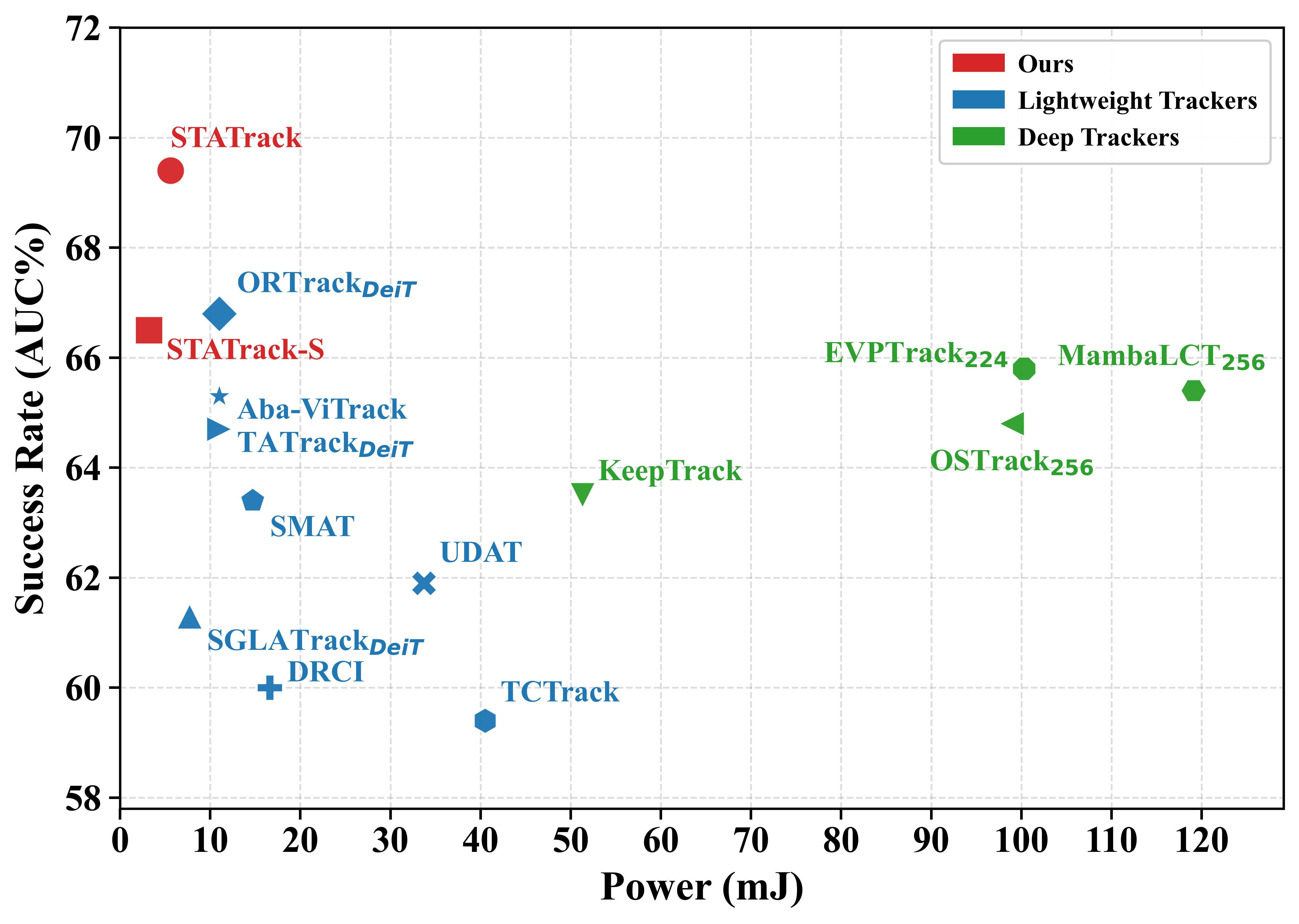}
    \caption{Accuracy--Energy Comparison on the VisDrone2018 dataset. STATrack achieves 69.4\% AUC with only 5.6 mJ theoretical inference energy, demonstrating a favorable accuracy--energy trade-off.}
	\label{FIG:power}
\end{figure}

To address the energy-efficiency bottleneck, spiking neural networks (SNNs) have recently attracted increasing attention because of their event-driven computational paradigm and sparse spike-based communication mechanism \cite{ESpikeFormer,SNNTrack}. On neuromorphic hardware, SNNs can exploit sparse spiking activity to convert some energy-intensive multiply-and-accumulate operations into energy-efficient accumulation operations, thereby offering considerable potential for low-energy inference \cite{SpikeFET,SDTrack}. Existing SNN-based trackers, such as SDTrack \cite{SDTrack} and SNNTrack \cite{SNNTrack}, are primarily designed for event-camera inputs and exploit the high temporal resolution and sparsity of event streams for target representation. More recently, SpikeFET \cite{SpikeFET} introduced a fully spiking tracking framework that fuses frame and event data. Compared with standard RGB cameras widely deployed on UAV platforms, event sensors introduce additional hardware costs, while event-based tracking datasets for UAV scenarios remain limited, thereby hindering the widespread deployment of these methods on existing UAV platforms. On the other hand, when processing standard RGB images with a limited number of time steps, discrete spiking representations still face challenges in preserving fine-grained target appearance information \cite{ESpikeFormer,IMLoss,RecDisSNN}. Therefore, how to achieve accuracy comparable to that of ANN-based trackers while retaining the energy-efficiency advantages of fully spiking computation still poses a significant challenge.  

To overcome these, we propose STATrack, a fully spiking UAV visual tracking framework that operates exclusively on RGB inputs. STATrack introduces an Adaptive Mutual Information Maximization (AMIM) mechanism to enhance deep spiking representation learning without altering the inference architecture. Specifically, AMIM maximizes the statistical dependency between target-centered template inputs and their deep spiking representations, encouraging the spiking backbone to learn discriminative representations that better distinguish targets from surrounding backgrounds. Considering that training samples exhibit different localization difficulties, we further develop a sample-difficulty-aware dynamic weighting strategy. This strategy adaptively adjusts the strength of the mutual information constraint according to the GIoU loss of the current batch and its historical exponential moving average. AMIM is used only during training and therefore introduces no additional computation during inference. 

Extensive experiments on four mainstream UAV tracking benchmarks demonstrate that STATrack achieves a favorable balance between tracking accuracy and energy efficiency. Specifically, STATrack obtains the best average success rate of 68.0\% and requires only 5.6 mJ theoretical energy consumption, achieving about $1.96\times$ better energy efficiency than ORTrack-DeiT. Moreover, STATrack-S further improves the energy efficiency to about $3.44\times$ while maintaining a competitive average success rate of 66.8\%. Moreover, as shown in Fig. \ref{FIG:power}, STATrack achieves the highest tracking accuracy on VisDrone2018 with the lowest theoretical energy consumption among the compared methods. It not only outperforms most lightweight trackers, but also surpasses representative deep trackers. These results demonstrate the potential of fully spiking architectures for energy-efficient UAV visual tracking. The main contributions of this work are summarized as follows:
\begin{itemize}
\item To the best of our knowledge, STATrack is the first fully spiking neural network framework specifically designed for UAV visual tracking using standard RGB images, achieving competitive tracking accuracy with low theoretical energy consumption.

\item We propose AMIM to learn more discriminative spiking representations by maximizing the dependency between target-centered templates and their deep features. A difficulty-aware weighting strategy adaptively regulates this constraint without introducing additional inference overhead.

\item Extensive experiments on four public UAV tracking benchmarks demonstrate that STATrack achieves state-of-the-art tracking performance with low theoretical energy consumption, highlighting its practical potential for energy-constrained UAV applications.
\end{itemize}

\section{Related Work}
\subsection{Efficient Visual Object Tracking}
In recent years, significant progress has been made in visual object tracking. The mainstream methods can be mainly divided into two categories: Discriminative Correlation Filter (DCF)-based methods and deep learning-based methods \cite{KCF,fDSST}. Early DCF trackers utilized fast Fourier transforms for efficient computation in the frequency domain, and due to their extremely high processing speed, they were favored in UAV tracking \cite{RACF}. However, DCF methods, relying on handcrafted features, often lack robustness when facing complex backgrounds, deformations, and occlusions. With the development of deep learning, CNN-based Siamese networks \cite{siamfc++,Spikingsiamfc++} and Transformer-based single-stream trackers \cite{OSTtrack,Mixformer} have gradually become dominant. With their powerful self-attention mechanism, Vits are capable of capturing long-range dependencies and unifying feature extraction and interaction processes, significantly improving tracking accuracy. However, these Transformer-based models typically come with large numbers of parameters and high computational overhead, making it difficult to achieve long-duration real-time deployment on UAV platforms with limited battery capacity and computing power \cite{AbaViTrack}. To achieve real-time tracking on UAVs, recent studies have proposed various lightweight strategies: Aba-ViTrack \cite{AbaViTrack} combines adaptive background-aware token computation with a lightweight ViT to achieve an excellent balance between accuracy and speed. AVTrack \cite{AVTrack} designs a hierarchical activation module that dynamically skips certain Transformer layers based on the input sample's complexity to accelerate inference. ORTrack \cite{ORTrackDeiT} introduces a mask-based occlusion simulation framework to learn robust representations and further improves tracking efficiency by incorporating adaptive feature-based knowledge distillation. Despite these methods alleviating computational pressure to some extent, achieving high-precision tracking under extreme low-power constraints remains challenging, prompting us to explore tracking methods with even lower energy consumption potential.

\subsection{Object Tracking Based on Spiking Neural Networks} 
To break through the energy-efficiency bottleneck of traditional ANNS, Spiking Neural Networks (SNNs), as third-generation neural networks, have gained widespread attention \cite{S4KD,ESpikeFormer}. SNNs communicate using discrete spike signals, transforming the high-energy multiply-accumulate operations (MACs) of ANNs into low-energy add-and-count operations (ACs), offering significant advantages on computationally constrained edge devices. In single-object tracking tasks, early frame-based SNN tracking methods \cite{S4KD,Spikingsiamfc++} typically employed Siamese network architectures with multi-step temporal encoding mechanisms to enhance temporal information modeling. However, these methods still lag behind ANNs in tracking accuracy and fail to fully exploit the energy efficiency potential of SNNs. With the development of event cameras, researchers have further explored event-driven SNN tracking schemes \cite{STNet,SCTN,SNNTrack}. For example, STNet \cite{STNet} utilizes SNNs to process high temporal resolution event streams, enhancing spatiotemporal feature extraction capability; SDTrack \cite{SDTrack} introduces a Transformer-based spiking-driven framework, achieving more efficient spatiotemporal modeling; while SpikeFET \cite{SpikeFET} proposes a fully spiking “frame-event” fusion architecture, leveraging multimodal complementarity to further improve tracking performance. Although these methods demonstrate superior spatiotemporal modeling capabilities in event-driven scenarios, their reliance on expensive event sensors limits their application in mainstream UAVs and embedded systems that are equipped only with RGB cameras. Currently, there is still a lack of an efficient SNN-based tracking framework that can achieve a favorable balance between tracking accuracy and energy efficiency using only RGB inputs, which is essential for practical deployment in UAV scenarios.
\begin{figure*}
	\centering
	\includegraphics[width=0.98\textwidth]{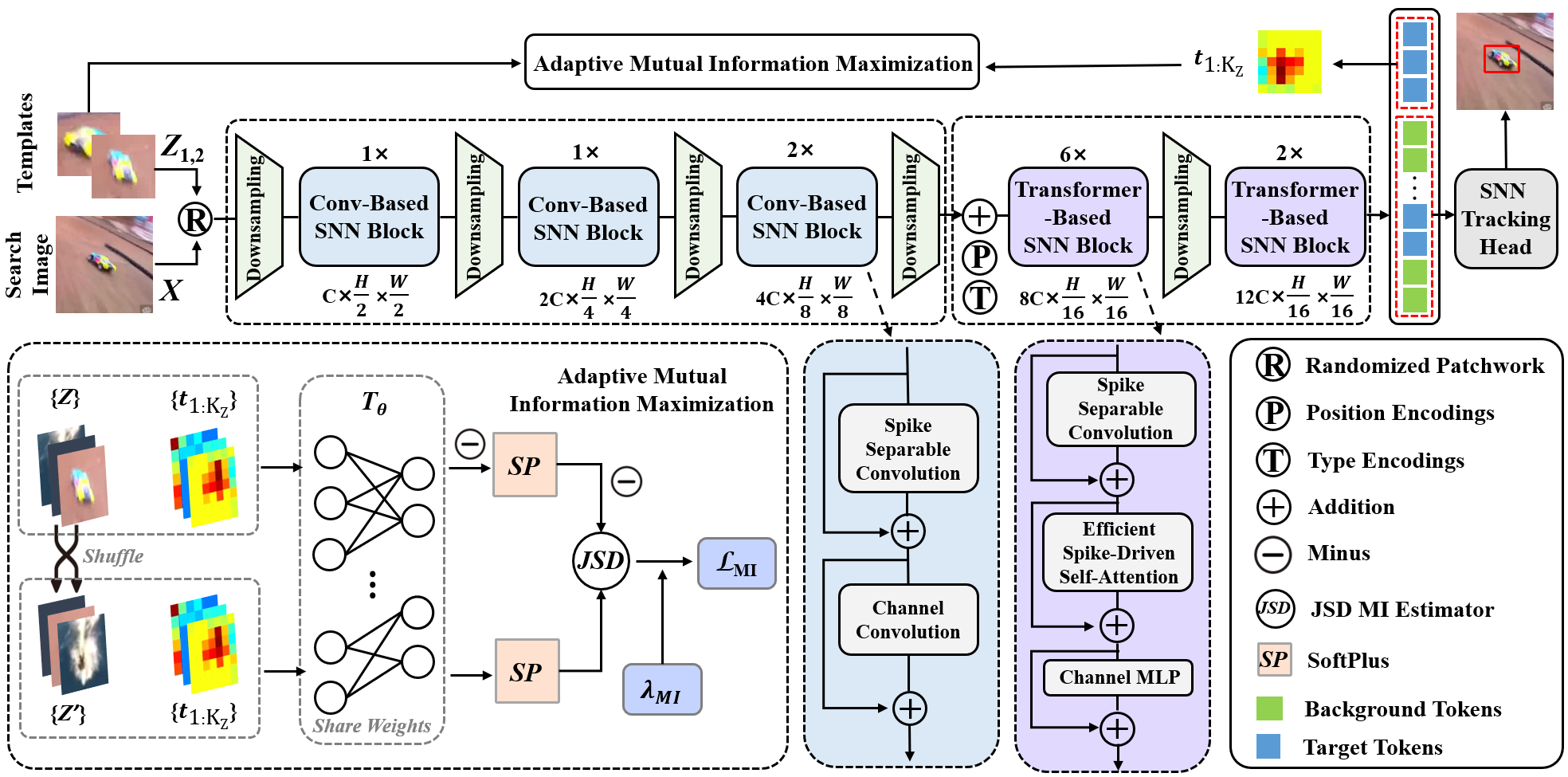}
	\caption{Overview of the proposed framework. The architecture consists of an efficient SNN-based Transformer backbone for joint template–search feature learning, followed by a spiking head. (Bottom-left) Illustration of the proposed Adaptive Mutual Information Maximization (AMIM) module. (Bottom-right) Detailed structure of the adaptive dynamic weighting strategy used in AMIM. Note that $\{Z\}$ denotes a batch of input samples, while $\{Z'\}$ represents a randomly shuffled batch of $Z$.}
	\label{FIG:pipeline}
\end{figure*}

\subsection{Target-Aware Feature Learning via Mutual Information}

Mutual information (MI) is a fundamental measure of the statistical dependency between random variables and has been widely applied to deep representation learning \cite{DeepInfoMax}. According to the InfoMax principle \cite{Selforganization}, maximizing the mutual information between an input and its latent representation can encourage deep representations to retain information associated with the input. In visual tasks, mutual information maximization has been applied to self-supervised learning, cross-modal representation learning, and target-aware representation learning \cite{Targetaware,Learningtarget}. In visual object tracking, the template generally contains concentrated target appearance information, whereas the search region includes substantial background content. Since the search region typically contains considerably more tokens than the template, jointly modeling the template and search region may gradually weaken target-related cues as features propagate through deep Transformer layers. To address this issue, TATrack \cite{TATrack} models the mutual information between the template input and its deep representation during training, encouraging the network to retain information associated with the target template and thereby enhancing target-aware representation learning. Existing mutual information-based target-aware methods are primarily designed for ANN or ViT trackers with continuous-valued feature representations, while their effectiveness in fully spiking tracking networks has not been sufficiently investigated. Compared with continuous-valued representations, discrete spiking representations operating with a limited number of time steps face additional challenges in preserving fine-grained target appearance information \cite{IMLoss,RecDisSNN}. This issue is particularly relevant to UAV tracking, where targets often occupy small image regions and are easily affected by complex backgrounds. Motivated by these observations, we apply a mutual information constraint to fully spiking UAV tracking by modeling the statistical dependency between target-centered template inputs and their deep spiking representations. This constraint encourages the network to learn discriminative representations that better distinguish targets from surrounding backgrounds. Unlike existing methods that employ a fixed mutual information weight, we further introduce a sample-difficulty-aware dynamic weighting strategy that adaptively adjusts the strength of the mutual information constraint according to the relative localization difficulty of the current batch. This constraint is applied only during training and therefore introduces no additional computation during inference.

\section{Proposed Approach}\label{sec:method}

This section presents the proposed STATrack framework for energy-efficient UAV visual tracking. We first provide an overview of the overall RGB-only fully spiking tracking architecture in Section \ref{sec:overview}. Then, the spiking neuron model used for efficient training and spike-driven inference is introduced in Section \ref{sec:neuron}. Section \ref{sec:backbone} describes the network architecture, including the Conv-Based SNN Block, the Transformer-Based SNN Block and prediction head. Section \ref{sec:amim} details the proposed Adaptive Mutual Information Maximization (AMIM) mechanism and its adaptive dynamic weighting strategy. Finally, Section \ref{sec:head_loss} introduces training objective and inference procedure.

\subsection{Overview}\label{sec:overview}

As illustrated in Fig.~\ref{FIG:pipeline}, STATrack is a fully spiking neural network framework designed for RGB-only UAV visual tracking. Given two template images $Z_1$ and $Z_2$ and a search image $X$, the Random Patch Module (RPM) \cite{SpikeFET} randomly arranges them through horizontal or vertical concatenation to construct a compact joint template and search input. This operation enables the three inputs to be processed within a unified feature extraction stream while alleviating the degradation of translation invariance caused by convolutional padding. The resulting joint RGB input is subsequently fed into the E-SpikeFormer \cite{ESpikeFormer} backbone for hierarchical spiking feature extraction. The backbone first employs Conv-Based SNN Blocks to capture local spatial structures and multi-scale visual patterns, followed by Transformer-Based SNN Blocks to model long-range dependencies between the template and search regions. Before entering the Transformer-Based SNN Blocks, positional and type encodings are incorporated to preserve spatial ocations and distinguish tokens originating from different input regions. The resulting deep spiking features are then separated according to the RPM arrangement, and the search region features are delivered to a lightweight fully spiking prediction head to estimate the target location. During training, we further introduce an Adaptive Mutual Information Maximization (AMIM) mechanism. AMIM models the statistical dependency between target-centered template inputs and their deep spiking representations, encouraging the spiking backbone to learn discriminative representations that better distinguish targets from surrounding backgrounds. Since AMIM serves only as an auxiliary training constraint and is removed during inference, it improves tracking accuracy without introducing additional inference computation.

\subsection{Spiking Neuron}\label{sec:neuron}

We employ the Spiking Firing Approximation (SFA) method \cite{ESpikeFormer,SpikeFET}. This approach aims to resolve the dilemma in SNN training where enforcing strict binary spikes causes severe gradient quantization errors, while employing multi-bit or continuous approximations compromises event-driven efficiency. Specifically, we postulate that spiking neurons adhere to a rate coding mechanism. Given a spike sequence at the $l$-th layer over $T$ discrete time steps, the average firing rate is defined as:
\begin{equation}
a^l = \frac{1}{T} \sum_{t=1}^{T} S^l[t],
\end{equation}
where $S^l[t] \in \{0, 1\}$ denotes the binary spike event at time step $t$. During the training phase, to mitigate the gradient quantization problem, we utilize integer activation to approximate the accumulated spike firing over the temporal dimension. The total number of spikes within $T$ time steps is approximated as a single-step integer activation value $S^l_T$:
\begin{equation}
S^l_T = \lfloor \operatorname{clip}(x^l, 0, T) \rceil,
\end{equation}
where $x^l$ represents the membrane potential input, and $T$ serves as both the time window length and the maximum firing threshold. This integer approximation maintains consistency with spiking behavior while enabling effective optimization via standard backpropagation.

During the inference phase, the model reverts to the spike-driven mode, recovering the activation value by accumulating binary spikes over $T$ time steps:
\begin{equation}
S^l_T = \sum_{t=1}^{T} \hat{S}^l[t],
\end{equation}
where $\hat{S}^l[t]$ is the actually generated spike sequence. Accordingly, the input for the subsequent layer is computed as:
\begin{equation}
X^{l+1} = W^{l+1} a^l = \frac{1}{T} W^{l+1} S^l_T.
\end{equation}
Since the spike sequence $\hat{S}^l[t]$ consists solely of binary events, all Multiply-Accumulate (MAC) operations are converted into sparse Accumulate (AC) operations, thereby supporting fully event-driven and energy-efficient inference.

\subsection{Network Architecture}\label{sec:backbone}

STATrack adopts E-SpikeFormer \cite{ESpikeFormer} as the spiking backbone, which consists of Conv-Based SNN Blocks and Transformer-Based SNN Blocks. The Conv-Based SNN Blocks are placed in the early stages to extract local spatial structures and multi-scale spiking representations, while the Transformer-Based SNN Blocks are used in deeper stages to model long-range dependencies between template and search tokens.

\noindent\textbf{Conv-Based SNN Block.}
Given an input feature $\mathbf{U}$, the Conv-Based SNN Block sequentially performs spatial token mixing and channel-wise feature interaction through residual spiking convolution operations:
\begin{align}
\mathbf{U}' &= \mathbf{U} + \mathrm{SSConv}(\mathbf{U}), \label{eq:U_prime}\\
\mathbf{U}'' &= \mathbf{U}' + \mathrm{ChConv}(\mathbf{U}'). \label{eq:U_double_prime}
\end{align}
where $\mathrm{SSConv}(\cdot)$ denotes the spike separable convolution module for spatial token mixing, and $\mathrm{ChConv}(\cdot)$ denotes the channel convolution module for channel-wise interaction. Specifically, they are formulated as:
\begin{align}
\mathrm{SSConv}(\mathbf{U})
&= \mathrm{Conv}_{\mathrm{pw}}
\big(\mathcal{SN}(
\mathrm{Conv}_{\mathrm{dw}}( \notag \\
&
\mathcal{SN}(
\mathrm{Conv}_{\mathrm{pw}}(\mathcal{SN}(\mathbf{U}))
)
)
)
\big), \\
\mathrm{ChConv}(\mathbf{U}')
&= \mathrm{Conv}
\big(
\mathcal{SN}(
\mathrm{Conv}(\mathcal{SN}(\mathbf{U}'))
)
\big).
\end{align}
Here, $\mathcal{SN}(\cdot)$ denotes the spiking neuron layer, $\mathrm{Conv}_{\mathrm{pw}}(\cdot)$ and $\mathrm{Conv}_{\mathrm{dw}}(\cdot)$ denote point-wise and depth-wise convolution, respectively. This block preserves local spatial details while maintaining sparse spike-driven computation.

\noindent\textbf{Transformer-Based SNN Block.}
To capture global template-search dependencies, the Transformer-Based SNN Block introduces spike-driven self-attention after local spiking convolution. Its computation is summarized as:
\begin{equation}
\begin{aligned}
\mathbf{U}' &= \mathbf{U} + \mathrm{SpikeSepConv}(\mathbf{U}), \\
\mathbf{U}'' &= \mathbf{U}' + \mathrm{SDSA}(\mathbf{U}'), \\
\mathbf{U}''' &= \mathbf{U}'' + \mathrm{ChannelMLP}(\mathbf{U}''),
\end{aligned}
\end{equation}
where $\mathrm{SDSA}(\cdot)$ denotes the efficient spike-driven self-attention module, and $\mathrm{ChannelMLP}(\cdot)$ denotes the channel feed-forward module. Different from conventional Transformer blocks that rely on dense continuous-valued attention, this block performs relation modeling in the spiking feature space, enabling STATrack to capture global dependencies while preserving event-driven computation.

\noindent\textbf{Spiking Prediction Head}
We adopt a lightweight spiking prediction head \cite{SpikeFET} composed of several Spiking Conv-BN-ReLU layers to directly regress the target bounding box while preserving the fully spiking and energy-efficient property of the network. Specifically, the output tokens corresponding to the search region are first reshaped into a 2D spatial feature map, which is then fed into the prediction head. The head produces three prediction branches: a classification score map $P \in [0,1]^{\frac{H_x}{s} \times \frac{W_x}{s}}$, a normalized bounding box size map $S \in [0,1]^{2 \times \frac{H_x}{s} \times \frac{W_x}{s}}$, and a local offset map $O \in [0,1]^{2 \times \frac{H_x}{s} \times \frac{W_x}{s}}$, 
where $s$ denotes the downsampling stride of the backbone. 
The preliminary estimation of the target position relies on identifying the location with the highest classification score, i.e., $(x_c, y_c) = \text{argmax}_{(x,y)} P(x, y)$. Subsequently, the final target bounding box is estimated by
\begin{equation}
\{(x_t, y_t); (w, h)\} = \{(x_c, y_c) + O(x_c, y_c); S(x_c, y_c)\}.
\end{equation}

\subsection{Adaptive Mutual Information Maximization (AMIM)}
\label{sec:amim}
To alleviate the issue that target-related information in deep spiking Transformers can be suppressed by dominant background tokens, we propose an Adaptive Mutual Information Maximization (AMIM) mechanism. As shown in Fig. \ref{FIG:pipeline}, AMIM employs a global statistics network to estimate the dependency between the template input $Z$ and its deep spiking feature representation $\mathbf{t}_{Z}$. By maximizing this dependency during training, AMIM encourages the backbone to preserve fine-grained and discriminative target semantics, thereby enhancing target-aware representation learning in cluttered UAV scenes.

\noindent\textbf{Target-aware Mutual Information Maximization.}
Mutual information (MI) quantifies the statistical dependency between two variables, making it suitable for preserving semantic consistency between an input image and its deep representation. However, directly computing MI is challenging due to the unknown underlying distributions in high-dimensional feature spaces. We adopt a Jensen--Shannon divergence (JSD)-based MI estimator to maximize the dependency between the template input and its deep spiking features. Specifically, the JSD-based MI estimate $\hat{I}^{(\mathrm{JSD})}$ is defined as:
\begin{equation}\label{eq:jsd}
\begin{split}
\hat{I}^{(\mathrm{JSD})}(Z, \mathbf{t}_{Z}; \theta) 
&= \mathbb{E}_{\mathbb{P}(Z, \mathbf{t}_{Z})} 
\left[ - \mathrm{sp} \left( -T_{\theta}(Z, \mathbf{t}_{Z}) \right) \right] \\
&\quad - \mathbb{E}_{\mathbb{P}(Z) \mathbb{P}(\mathbf{t}_{Z})} 
\left[ \mathrm{sp} \left( T_{\theta}(Z', \mathbf{t}_{Z}) \right) \right],
\end{split}
\end{equation}
where $T_{\theta}$ denotes a global statistics network parameterized by $\theta$, and $\mathrm{sp}(z)=\log(1+e^z)$ is the SoftPlus function. Here, $Z$ denotes a batch of template inputs, while $Z'$ represents a randomly shuffled version of $Z$. The shuffling operation makes $Z'$ statistically independent of $\mathbf{t}_{Z}$, thereby approximating samples from the product of marginal distributions. The detailed architecture of $T_{\theta}$ is provided in the supplementary material.

The objective of AMIM is to maximize the above MI estimate. Therefore, the MI loss is formulated as:
\begin{equation}\label{eq:mi_loss}
\mathcal{L}_{\mathrm{MI}} 
= - \lambda_{\mathrm{MI}} \cdot 
\hat{I}^{(\mathrm{JSD})}(Z, \mathbf{t}_{Z}; \theta),
\end{equation}
where $\lambda_{\mathrm{MI}}$ controls the strength of the MI constraint and is dynamically adjusted by the adaptive weighting strategy described below.

\noindent\textbf{Adaptive Dynamic Weighting (ADW).}
A fixed MI weight may not be optimal for all training samples, since UAV tracking sequences contain samples with varying degrees of difficulty, such as background clutter, occlusion, and motion blur. To stabilize MI-enhanced learning, we introduce an Adaptive Dynamic Weighting (ADW) strategy that adjusts $\lambda_{\mathrm{MI}}$ according to the relative difficulty of the current batch. Specifically, we use the exponential moving average (EMA) of the GIoU loss as a historical reference and compute the difficulty-aware adjustment term as:
\begin{equation}\label{eq:delta}
\Delta =
\tanh \left(
\eta \cdot
\left(
\bar{\mathcal{L}}_{\mathrm{giou}}
-
\mathcal{L}_{\mathrm{giou}}
\right)
\right),
\end{equation}
where $\bar{\mathcal{L}}_{\mathrm{giou}}$ denotes the historical EMA of the GIoU loss, $\mathcal{L}_{\mathrm{giou}}$ represents the GIoU loss of the current batch, and $\eta$ controls the sensitivity of the adjustment. The adaptive MI weight is then defined as:
\begin{equation}\label{eq:lambda}
\lambda_{\mathrm{MI}} =
\mathrm{Clamp}
\left(
\lambda_{\mathrm{base}} - \beta \cdot \Delta,
0,
\lambda_{\max}
\right),
\end{equation}
where $\lambda_{\mathrm{base}}$ is the base MI weight, $\beta$ controls the adjustment amplitude, and $\lambda_{\max}$ prevents excessively large MI regularization. When the current batch is more difficult than the historical average, i.e., $\mathcal{L}_{\mathrm{giou}} > \bar{\mathcal{L}}_{\mathrm{giou}}$, we have $\Delta < 0$, which increases $\lambda_{\mathrm{MI}}$ and strengthens target-aware semantic learning. Conversely, for easier samples, ADW reduces the MI constraint to avoid unnecessary regularization.
\subsection{Training and Inference}\label{sec:head_loss}

\noindent\textbf{Training.} We adopt a standard composite loss formulation. Specifically, a weighted focal loss $\mathcal{L}_{\mathrm{cls}}$ \cite{Focalloss} is used for classification, while bounding box regression is supervised by a combination of $L_1$ loss and GIoU loss \cite{Giou}. In addition, a similarity loss $\mathcal{L}_{\mathrm{sim}}$ is employed to maintain spatio-temporal consistency \cite{SpikeFET}.
The overall training objective is defined as:
\begin{equation}
\begin{split}
\mathcal{L}_{\mathrm{total}} &= \mathcal{L}_{\mathrm{cls}} +
\lambda_{\mathrm{giou}} \mathcal{L}_{\mathrm{giou}} +
\lambda_{L1} \mathcal{L}_{L1}  \\
& +
\lambda_{\mathrm{sim}} \mathcal{L}_{\mathrm{sim}} +
 \mathcal{L}_{\mathrm{MI}}.
\end{split}
\end{equation}
where $\lambda_{\mathrm{cls}}$, $\lambda_{\mathrm{giou}}$, 
$\lambda_{1}$, and $\lambda_{\mathrm{sim}}$ denote the weights of the focal loss, GIoU loss, $L_1$ loss, and similarity loss, respectively. 
They are set to $1$, $2$, $5$, and $0.5$, respectively. 
The mutual information loss is adaptively weighted by $\lambda_{\mathrm{MI}}$ through the proposed ADW strategy.

\noindent\textbf{Inference.} Since the Random Patch Module (RPM) takes two template images and one search region as input, STATrack maintains a two-slot template list during inference. The first slot stores the initial template and remains fixed throughout tracking, serving as a reliable identity anchor. The second slot is dynamically updated to capture recent target appearance variations. Specifically, every $T_u$ frames, if the confidence score of the current prediction is higher than a threshold $\tau$, a new template patch is cropped from the current frame according to the predicted target location and used to replace the second template slot. The two templates are then jointly fed into RPM together with the current search region. This design enables STATrack to preserve long-term target identity while adapting to short-term appearance changes, which is particularly useful for UAV tracking under viewpoint variation, scale change, and motion blur. In our implementation, we set $T_u=25$ and $\tau=0.7$.

\section{Experiments}
\label{section_experiment}
In this section, we conduct a comprehensive evaluation of the proposed STATrack on four widely used UAV tracking benchmarks, including UAV123 \cite{UAV123}, VisDrone2018 \cite{Visdrone2018}, UAVDT \cite{UAVDT}, and UAVTrack112 \cite{uavtrack112}. All experiments were conducted on a PC equipped with an Intel i9-10850K CPU (3.6 GHz), 16 GB RAM, and an NVIDIA Titan X GPU. To validate the effectiveness of STATrack in achieving competitive tracking performance with low energy consumption, we compare it with 19 state-of-the-art (SOTA) lightweight trackers under the same evaluation protocol. We separate them into three groups for a clearer comparison: 1) DCF-based trackers \cite{KCF,fDSST,ECO,RACF,AutoTrack}, 2) CNN-based trackers \cite{HiFT,TCTrack,SGDViT,DRCL,PRLTRack}, and 3) ViT-based trackers \cite{AbaViTrack,SMAT,TATrack,AVTrack,ORTrackDeiT,SGLATrackDeiT}.

\begin{table*}[t]
\centering
\caption{Comparison of STATrack with lightweight trackers in terms of precision (Prec.), success rate (Succ.), and energy consumption (mJ) on UAVTrack112, UAVDT, VisDrone2018, and UAV123. \textcolor{red}{\textbf{Red}}, \textcolor{blue}{\textbf{blue}}, and \textcolor{green}{\textbf{green}} indicate the first, second, and third best results, respectively. Note that percentage symbols (\%) are omitted for both precision and success rate values.}
\label{tab:sota}
\begin{tabular}{@{}cc cc cc cc cc cc cc cc@{}} 
\toprule
\multirow{2}{*}{\textbf{Tracker}} & \multirow{2}{*}{\textbf{Source}} & \multicolumn{2}{c}{\textbf{UAVTrack112}} & \multicolumn{2}{c}{\textbf{UAVDT}} & \multicolumn{2}{c}{\textbf{VisDrone2018}} & \multicolumn{2}{c}{\textbf{UAV123}} & \multicolumn{2}{c}{\textbf{Average}} & \textbf{Power} & \textbf{Param.} \\
\cmidrule(lr){3-4} \cmidrule(lr){5-6} \cmidrule(lr){7-8} \cmidrule(lr){9-10} \cmidrule(lr){11-12} 
& & Prec. & Succ. & Prec. & Succ. & Prec. & Succ. & Prec. & Succ. & Prec. & Succ. & \textbf{(mJ)} & \textbf{(M)} \\
\midrule
KCF \cite{KCF} & TPAMI 15 & 40.2 & 26.6 & 57.1 & 29.0 & 68.5 & 41.3 & 52.3 & 33.1 & 54.5 & 32.5 & - & - \\
fDSST \cite{fDSST} & TPAMI 17 & 56.8 & 39.1 & 66.6 & 38.3 & 69.8 & 51.0 & 58.3 & 40.5 & 62.9 & 42.2 & - & - \\
ECO\_HC \cite{ECO} & CVPR 17 & 68.6 & 47.4 & 69.4 & 41.6 & 80.8 & 58.1 & 71.0 & 49.6 & 72.5 & 49.2 & - & - \\
MCCT\_H \cite{MCCT} & CVPR 18 & 63.4 & 43.6 & 66.8 & 40.2 & 80.3 & 56.7 & 65.9 & 45.7 & 69.1 & 46.6 & - & - \\
ARCF \cite{ARCF} & ICCV 19 & 67.2 & 45.6 & 72.0 & 45.8 & 79.7 & 58.4 & 67.1 & 46.8 & 71.5 & 49.2 & - & - \\
AutoTrack \cite{AutoTrack} & CVPR 20 & 69.4 & 46.5 & 71.8 & 45.0 & 78.8 & 57.3 & 68.9 & 47.2 & 72.2 & 49.0 & - & - \\
RACF \cite{RACF} & PR 22 & 70.2 & 48.5 & 77.3 & 49.4 & 83.4 & 60.0 & 70.2 & 47.7 & 75.3 & 51.4 & -  & - \\ 
\midrule
HiFT \cite{HiFT} & ICCV 21 & 74.2 & 57.0 & 65.2 & 47.5 & 71.9 & 52.6 & 78.7 & 59.0 & 72.5 & 54.0 & 33.1 & 9.9 \\
TCTrack \cite{TCTrack} & CVPR 22 & 76.6 & 59.4 & 72.5 & 53.0 & 79.9 & 59.4 & 80.0 & 60.5 & 77.3 & 58.1 & 40.5 & 9.7 \\
UDAT \cite{UDAT} & CVPR 22 & 78.9 & 61.6 & 80.1 & 59.2 & 81.6 & 61.9 & 76.1 & 59.0 & 79.2 & 60.4 & 33.7 & - \\       
SGDViT \cite{SGDViT} & ICRA 23 & 73.6 & 59.9 & 65.7 & 48.0 & 72.1 & 52.1 & 75.4 & 57.5 & 71.7 & 54.4 & 60.0 & 23.3 \\
DRCI \cite{DRCL} & ICME 23 & 80.1 & 61.3 & \textcolor{blue}{\textbf{84.0}} & 59.0 & 83.4 & 60.0 & 76.7 & 59.7 & 81.1 & 60.0 & 16.6 & 8.8 \\
PRL-Track \cite{PRLTRack} & IROS 24 & 76.1 & 62.0 & 73.1 & 53.5 & 72.6 & 53.8 & 79.1 & 59.3 & 75.2 & 57.2 & 34.0 & 12.0 \\
\midrule
Aba-ViTrack \cite{AbaViTrack} & ICCV 23 & 82.6 &\textcolor{green}{\textbf{67.6}} & \textcolor{green}{\textbf{83.4}} & 59.9 & 86.1 & 65.3 & \textcolor{blue}{\textbf{86.4}} & 66.4 & \textcolor{green}{\textbf{84.6}} & 64.8 & 11.0 & 8.0 \\
SMAT \cite{SMAT} & WACV 24 & 82.2 & 65.3 & 80.8 & 58.7 & 82.5 & 63.4 & 81.8 & 64.6 & 81.8 & 63.0 & 14.7 & 8.6 \\
TATrack-DeiT \cite{TATrack} & TGRS 24 & 81.5 & 65.8 & \textcolor{green}{\textbf{83.4}} & \textcolor{green}{\textbf{60.6}} & 85.2 & 64.7 & 82.7 & 65.4 & 83.2 & 64.1 & 11.0 & 5.6 \\
AVTrack-DeiT \cite{AVTrack} & ICML 24 & 80.3 & 65.4 & 82.1 & 58.7 & 86.0 & 65.3 & 84.8 & \textcolor{green}{\textbf{66.8}} & 83.3 & 64.1 & 4.5-8.8 & 3.5-7.9 \\
ORTrack-Deit \cite{ORTrackDeiT} & CVPR 25 & \textcolor{red}{\textbf{83.2}} & 67.0 & \textcolor{green}{\textbf{83.4}} & 60.1 & \textcolor{blue}{\textbf{88.6}} & \textcolor{blue}{\textbf{66.8}} & 84.3 & 66.4 & \textcolor{blue}{\textbf{84.9}} & \textcolor{green}{\textbf{65.1}} & 11.0 & 5.8 \\
SGLATrack-Deit \cite{SGLATrackDeiT} & CVPR 25 & \textcolor{green}{\textbf{82.8}} &  67.5 & 81.9 & 59.9 & 80.0 & 61.3 & \textcolor{green}{\textbf{84.9}} & \textcolor{blue}{\textbf{66.9}} & 82.4 & 63.9 & 7.7 & 7.9 \\
\midrule
\textbf{STATrack} & \multirow{2}{*}{\textbf{Ours}} 
& \textcolor{blue}{\textbf{83.0}} & \textcolor{red}{\textbf{69.7}} 
& \textcolor{blue}{\textbf{84.0}} & \textcolor{blue}{\textbf{65.3}} 
& \textcolor{red}{\textbf{89.1}} & \textcolor{red}{\textbf{69.4}} 
& \textcolor{red}{\textbf{86.7}} & \textcolor{red}{\textbf{67.6}} 
& \textcolor{red}{\textbf{85.7}} & \textcolor{red}{\textbf{68.0}}
& \textbf{5.6} & \textbf{12.6} \\
\textbf{STATrack-S} &  
& 82.3 & \textcolor{blue}{\textbf{69.0}} 
& \textcolor{red}{\textbf{84.2}} & \textcolor{red}{\textbf{65.8}} 
& \textcolor{green}{\textbf{86.8}} & \textcolor{green}{\textbf{66.5}} 
& 84.2 & 65.8 
& 84.4 & \textcolor{blue}{\textbf{66.8}}
& \textbf{3.2} & \textbf{7.4} \\
\bottomrule
\end{tabular}
\end{table*}

\subsection{Implementation Details}

\noindent\textbf{Model Configuration}. 
The proposed STATrack adopts an efficient spike-driven Transformer backbone based on E-SpikeFormer \cite{ESpikeFormer}. Specifically, we adopt E-SpikeFormer-10M as the backbone for STATrack and E-SpikeFormer-5.1M as the backbone for STATrack-S. The prediction module is implemented as a lightweight spike-driven center prediction head, consisting entirely of spiking operations. For each of the three output branches, i.e., classification, bounding box size regression, and center offset regression, we use four stacked Multi-Spike convolutional layers, forming an end-to-end fully spiking processing pipeline. Following the standard settings of Transformer-based trackers \cite{OSTtrack,Mixformer}, the input resolutions of the template and search region are set to $128 \times 128$ and $256 \times 256$, respectively. The time step of spiking neurons is set to $T=4$ in all experiments. Integer activations within $[0,T]$ are used during training, while the network is unfolded into $T$ binary spiking timesteps during inference to preserve spike-driven computation.

\noindent\textbf{Training Strategy}. 
STATrack is trained on a combination of large-scale tracking datasets, including GOT-10k \cite{got10k}, LaSOT \cite{Lasot}, COCO \cite{coco}, and TrackingNet \cite{Trackingnet}. To ensure fair comparisons, the training pipeline follows the same protocol as the baseline methods. The batch size is set to 32, and the network is optimized using the AdamW optimizer with a weight decay of $1 \times 10^{-4}$. The initial learning rate is set to $6 \times 10^{-5}$. The model is trained for 300 epochs with 60,000 image pairs per epoch, and the learning rate is decayed by a factor of 10 after the 240th epoch. For the proposed Adaptive Mutual Information Maximization (AMIM) loss, the base weight $\lambda_{\mathrm{base}}$, dynamic range $\beta$, and sensitivity coefficient $\eta$ are empirically set to 0.1, 0.05, and 10.0, respectively.

\begin{figure*}[t]
    \centering
    \includegraphics[width=0.49\textwidth]{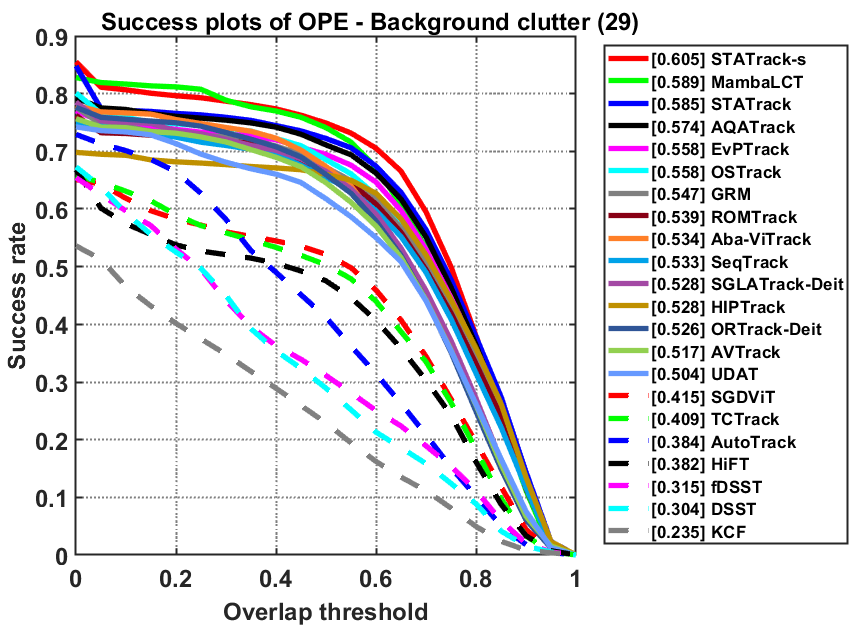}
    \hfill
    \includegraphics[width=0.49\textwidth]{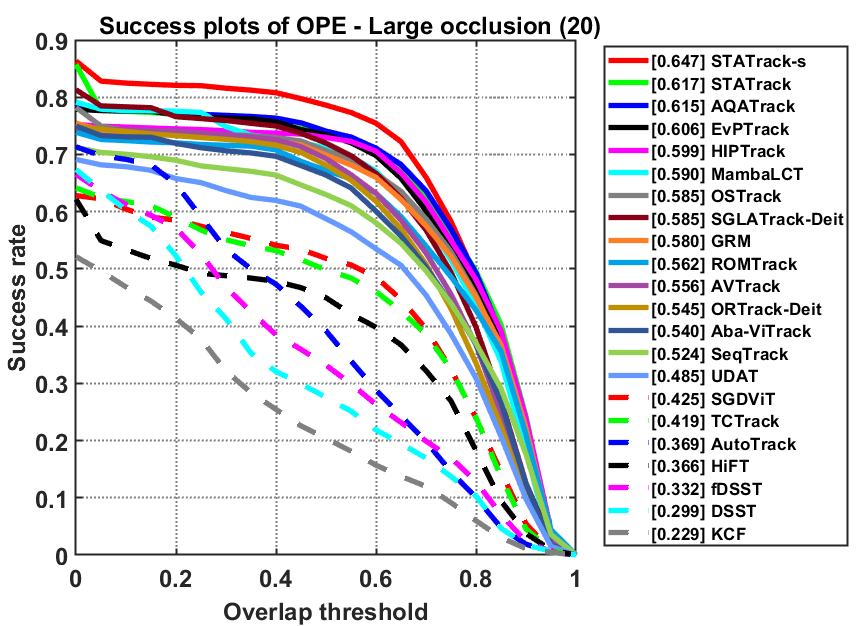}
    \caption{Attribute-based success plots on the UAVDT dataset under two representative challenging attributes: background clutter and large occlusion.}
    \label{fig:attribute_uavdt}
\end{figure*} 
\noindent\textbf{Inference Procedure}. 
During inference, we follow the standard tracking practice by applying a Hanning window penalty to introduce motion smoothness and spatial priors. Specifically, the predicted center classification score map is multiplied element-wise by a Hanning window of the same resolution, and the bounding box corresponding to the highest response is selected as the final tracking result.

\begin{table*}[t]
\centering
\caption{Comparison results of STATrack with deep learning-based trackers on the VisDrone2018 dataset.}
\label{tab:visdrone_bmtrack}

\begin{tabular}{lccc|lccc|lccc}
\hline
Method & Source & Prec. & Succ. &
Method & Source & Prec. & Succ. &
Method & Source & Prec. & Succ.  \\
\hline

\textbf{STATrack} & \textbf{Ours} &  \textcolor{red}{\textbf{89.1}} &  \textcolor{red}{\textbf{69.4}} &
EVPTrack \cite{EvpTrack} & AAAI 24 & 84.5 & 65.8 &
SparseTT \cite{SparseTT}  & IJCAI 22 & 81.4 & 62.1 \\

MambaLCT \cite{Mambalct} & AAAI 25 &  \textcolor{blue}{\textbf{88.1}} & 65.5 &
ZoomTrack \cite{ZoomTrack}  & NIPS 23 & 81.4 & 63.4 &
OSTrack \cite{OSTtrack} & ECCV 22 & 84.2 & 64.8 \\

HIPTrack \cite{Hiptrack} & AAAI 24 & 86.7 &  \textcolor{blue}{\textbf{67.1}} &
SeqTrack \cite{Seqtrack} & CVPR 23 & 85.3 & 65.8 &
SAOT \cite{SOTA} & ICCV 21 & 76.9 & 59.1 \\

AQATrack \cite{AQATrack}  & CVPR 24 & \textcolor{green}{\textbf{87.2}} &  \textcolor{green}{\textbf{66.9}} &
MAT\cite{MAT} & CVPR 23 & 81.6 & 62.2 &

KeepTrack \cite{KeepTrack} & ICCV 21 & 84.0 & 63.5 \\
ODTrack \cite{ODTrack} & AAAI 24 & 85.6 & 64.8 &
ROMTrack \cite{ROMTrack} & ICCV 23 & 86.4 & 66.7 &
PrDiMP50 \cite{PrDiMP} & CVPR 20 & 79.4 & 59.7 \\
\hline
\end{tabular}

\end{table*}

\noindent\textbf{Energy Consumption.}
To evaluate energy efficiency, we adopt the commonly used theoretical energy estimation protocol for SNNs. Under a 45\,nm CMOS technology, the energy costs of one MAC operation and one AC operation are set to $E_{\mathrm{MAC}}=4.6$\,pJ and $E_{\mathrm{AC}}=0.9$\,pJ, respectively \cite{Cmos,metaformerv2}. For ANN-based trackers, energy consumption is estimated by the number of dense MAC operations:
\begin{equation}
E_{\mathrm{ANN}} = E_{\mathrm{MAC}} \cdot \mathrm{FL}_{\mathrm{ANN}},
\end{equation}
where $\mathrm{FL}_{\mathrm{ANN}}$ denotes the total FLOPs of the ANN model.

For SNN-based trackers, spike-driven layers mainly rely on sparse AC operations. Therefore, the energy consumption is estimated by considering the temporal length $T$, layer-wise firing rates, and operation counts:
\begin{equation}
\begin{aligned}
E_{\mathrm{SNN}} =
&T \cdot E_{\mathrm{AC}}
\left(
\sum_{m=1}^{M} R_C^{(m)} \mathrm{FL}_{\mathrm{Conv}}^{(m)}
+
\sum_{n=1}^{N} R_M^{(n)} \mathrm{FL}_{\mathrm{MLP}}^{(n)}
\right) \\
&+ T \cdot E_{\mathrm{MAC}} \cdot \mathrm{FL}_{\mathrm{head}},
\end{aligned}
\end{equation}
where $R_C^{(m)}$ and $R_M^{(n)}$ denote the firing rates of the $m$-th convolutional layer and the $n$-th MLP layer, respectively. The firing rate is defined as the ratio of non-zero elements in the corresponding spike tensor. $\mathrm{FL}_{\mathrm{head}}$ denotes the FLOPs of non-spiking convolutional layers in the tracking head.

The FLOPs of convolutional and MLP layers are computed as:
\begin{align}
\mathrm{FL}_{\mathrm{Conv}}^{(m)}
&=
k_m^2 \cdot h_m \cdot w_m \cdot c_{m-1} \cdot C_m, \label{eq:flops_conv}\\
\mathrm{FL}_{\mathrm{MLP}}^{(n)} 
&=
i_n \cdot o_n. \label{eq:flops_mlp}
\end{align}
where $k_m$ is the kernel size, $(h_m,w_m)$ is the output feature map size, $c_{m-1}$ and $C_m$ are the input and output channels, and $i_n$ and $o_n$ are the input and output dimensions of the MLP layer. See the supplementary material for details.

\subsection{State-of-the-art Comparison}

\noindent\textbf{Comparison with Lightweight Trackers} We compare the proposed STATrack with 19 state-of-the-art lightweight UAV trackers on four challenging benchmarks: UAVTrack112, UAVDT, VisDrone2018, and UAV123. The overall comparison results are reported in Table \ref{tab:sota}, covering representative trackers based on DCF, CNN, and ViT architectures.

Compared with DCF-based trackers, STATrack consistently achieves substantial performance improvements across the four benchmarks. Specifically, STATrack obtains an average precision of 85.7\% and an average success rate of 68.0\%, outperforming the best-performing DCF-based tracker RACF \cite{RACF} by 10.4\% and 16.6\%, respectively. These results demonstrate the clear performance advantage of the proposed fully spiking framework over conventional DCF-based UAV trackers.

STATrack also shows clear advantages over CNN-based lightweight trackers in terms of both tracking accuracy and theoretical energy consumption. Its average precision improves upon DRCI \cite{DRCL} from 81.1\% to 85.7\%, corresponding to a gain of 4.6\%. Its average success rate also exceeds the strongest CNN-based result achieved by UDAT \cite{UDAT}, increasing from 60.4\% to 68.0\%, a gain of 7.6\%. Moreover, STATrack requires only 5.6 mJ of theoretical energy consumption, corresponding to approximately 33.7\% of that required by DRCI. These results indicate that STATrack provides a more favorable balance between tracking accuracy and energy consumption than the compared CNN-based UAV trackers.

\begin{figure*}[t]
    \centering
    \includegraphics[width=1.0 \textwidth]{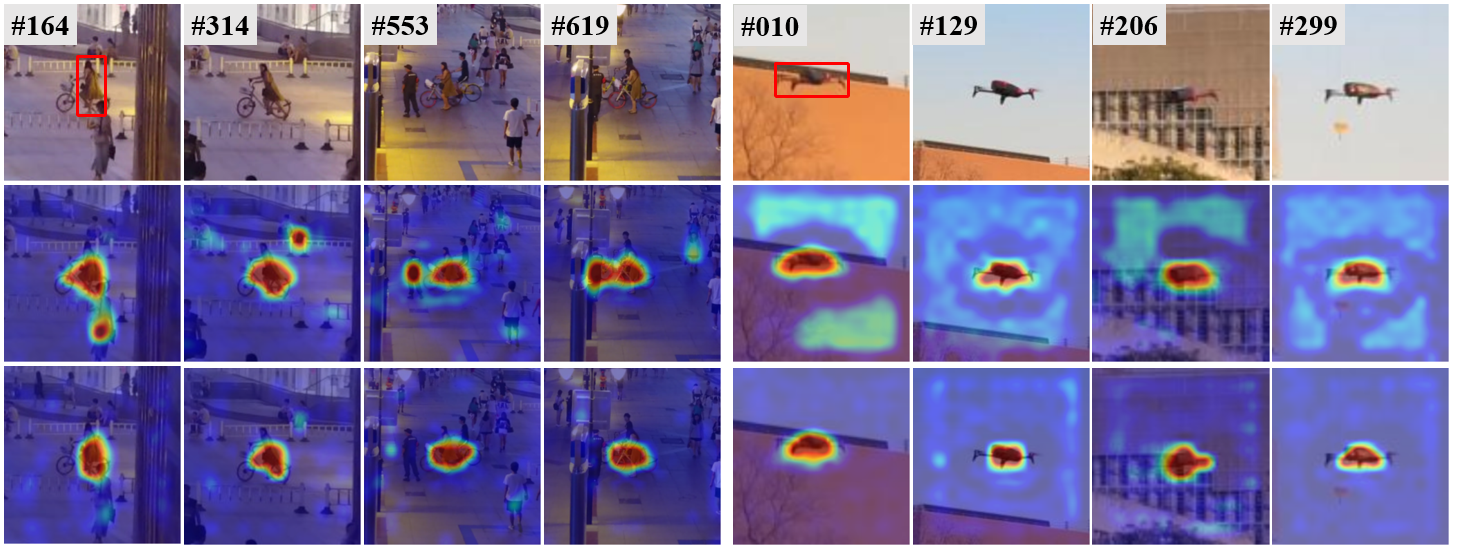}
    \caption{Qualitative comparison of attention maps with and without AMIM. The top row shows the search images, while the middle and bottom rows show the attention maps generated by the baseline without AMIM and STATrack with AMIM, respectively.} 
\label{fig:feature}
\end{figure*}
\begin{table*}[t]
\centering
\caption{Evaluation of CPU and GPU Speed for different methods.}
\label{tab:speed}
\begin{tabular}{lccccccc}
\toprule
\textbf{Method} 
& \begin{tabular}{c}STATrack\\(Ours) \end{tabular}
& \begin{tabular}{c}STATrack-S\\(Ours) \end{tabular}
& \begin{tabular}{c}ORTrack\\-DeiT \cite{ORTrackDeiT}\end{tabular}
& \begin{tabular}{c}AVTrack\\-DeiT \cite{AVTrack}\end{tabular}
& \begin{tabular}{c}OSTrack\\-256 \cite{OSTtrack}\end{tabular}
& \begin{tabular}{c}MambaLCT\\-256 \cite{Mambalct}\end{tabular}
& \begin{tabular}{c}EVPTrack\\-224 \cite{EvpTrack}\end{tabular} \\
\midrule
\textbf{Power (mJ)} & 5.6 & 3.2 & 11.0 & 4.5--8.8 & 98.9 & 119.1 & 100.3 \\
\textbf{FPS (GPU)}     & 118.4 & 156.7 & 226.4 & 260.3 & 65.4 & 57.4 & 74.3 \\
\textbf{FPS (CPU)}     & 24.2 & 32.1 & 55.4 & 59.8 & 5.9 & -- & 6.7 \\
\bottomrule
\end{tabular}
\end{table*}
Compared with recent ViT-based trackers, STATrack achieves superior average tracking performance while maintaining low theoretical energy consumption. STATrack obtains the highest average precision of 85.7\%, exceeding ORTrack-DeiT \cite{ORTrackDeiT}, the strongest competing method, by 0.80\%. It also achieves the highest average success rate of 68.0\%, outperforming ORTrack-DeiT, Aba-ViTrack \cite{AbaViTrack}, TATrack-DeiT \cite{TATrack}, AVTrack-DeiT \cite{AVTrack}, and SGLATrack-DeiT \cite{SGLATrackDeiT}. In particular, STATrack achieves the best success rates on UAVTrack112, VisDrone2018, and UAV123, reaching 69.7\%, 69.4\%, and 67.6\%, respectively. It also obtains a success rate of 65.3\% on UAVDT, ranking second only to STATrack-S. These results demonstrate that STATrack consistently delivers strong tracking performance across diverse UAV benchmarks.

In terms of efficiency, STATrack benefits from sparse spike-driven computation and requires only 5.6 mJ of theoretical energy consumption, which is lower than that of most ANN-based lightweight trackers in the comparison. The compact STATrack-S further reduces the theoretical energy consumption to 3.2 mJ while achieving an average success rate of 66.8\%, ranking second among all compared methods. Compared with STATrack and ORTrack-DeiT, STATrack-S reduces the theoretical energy consumption by 42.9\% and 70.9\%, respectively. Despite its substantially lower energy requirement, STATrack-S surpasses ORTrack-DeiT by 1.7\% in average success rate. These results demonstrate that the proposed fully spiking framework provides a favorable balance between tracking accuracy and theoretical energy consumption for resource-constrained UAV visual tracking.

\noindent\textbf{Comparison with Deep Trackers.}
We further compare STATrack with 14 recent deep learning-based trackers on the VisDrone2018 benchmark. As shown in Table \ref{tab:visdrone_bmtrack}, STATrack achieves the best precision and success rate, reaching 89.1\% and 69.4\%, respectively. Compared with MambaLCT, the strongest competing method in precision, STATrack improves the precision by 1.0\% and the success rate by 3.9\%. Moreover, STATrack outperforms strong Transformer-based trackers, including HIPTrack, AQATrack, and ROMTrack, in both metrics.

\noindent\textbf{Attribute-Based Evaluation.}
To verify the robustness of STATrack under typical UAV tracking challenges, we compare STATrack and STATrack-S with 20 advanced trackers on two representative attribute subsets, as shown in Fig. \ref{fig:attribute_uavdt}. Under background clutter, STATrack-S achieves the highest success rate of 60.5\%, while STATrack ranks third with 58.5\%, only 0.4\% lower than the deep tracker MambaLCT. This result verifies the effectiveness of AMIM in reducing background interference and enhancing target-aware representations. Under large occlusion, STATrack-S and STATrack achieve the top two results, with success rates of 64.7\% and 61.7\%, respectively, indicating that the proposed framework can preserve discriminative target semantics under severe occlusion.

\noindent\textbf{Speed Evaluation.}
For a fair speed comparison, Table \ref{tab:speed} reports the CPU/GPU inference speed of STATrack and other representative trackers measured on the same hardware platform. STATrack achieves 118.4 FPS on GPU and 24.2 FPS on CPU, while STATrack-S reaches 156.7 FPS on GPU and 32.1 FPS on CPU. These results indicate that the proposed tracker can achieve real-time or near real-time performance even on conventional hardware without specialized neuromorphic acceleration. It should be noted that traditional CPU/GPU platforms are mainly optimized for dense MAC operations rather than sparse spike-driven AC operations. Therefore, these FPS results should be regarded as conservative runtime references rather than the upper-bound efficiency of SNN deployment. On neuromorphic or spike-oriented hardware, STATrack may further benefit from its event-driven computation pattern.

\begin{table*}[t]
\centering
\small 
\caption{Ablation results of different methods in STATrack on four UAV benchmarks. \textbf{MIM} refers to mutual information maximization with a fixed weighting scheme, while \textbf{ADW} denotes the proposed adaptive weighting strategy. Best results are shown in \textbf{bold}.}
\label{tab:ablation}
\begin{tabular}{@{}cccccccccccc@{}}
\toprule
\multicolumn{2}{c}{\textbf{Components}} & \multicolumn{2}{c}{\textbf{UAVTrack112}}  & \multicolumn{2}{c}{\textbf{UAVDT}} & \multicolumn{2}{c}{\textbf{VisDrone2018}} & \multicolumn{2}{c}{\textbf{UAV123}} & \multicolumn{2}{c}{\textbf{Average}} \\
\cmidrule(r){1-2} \cmidrule(lr){3-4} \cmidrule(lr){5-6} \cmidrule(lr){7-8} \cmidrule(lr){9-10} \cmidrule(l){11-12} 
MIM & ADW & Prec. & Succ. & Prec. & Succ. & Prec. & Succ. & Prec. & Succ. & Prec. & Succ. \\
\midrule
- & -  & 80.3 & 67.7 & 78.5 & 61.2 & 82.5 & 65.0 & 86.1 & 67.0 & 81.9 & 65.2 \\
\checkmark & - & 81.8 & 68.2 & 81.5 & 63.4 & 85.6 & 67.6 & 86.5 & 67.3 & 83.9 & 66.6 \\
\checkmark & \checkmark & \textbf{83.0} & \textbf{69.7} & \textbf{84.0} & \textbf{65.3} 
& \textbf{89.1} & \textbf{69.4} & \textbf{86.7} &\textbf{67.6} & \textbf{85.7} & \textbf{68.0} \\
\bottomrule
\end{tabular}
\end{table*}

\begin{table*}[t]
\centering
\caption{Parameter sensitivity analysis on four benchmarks. We investigate the impact of the base weight $\lambda_{\mathrm{base}}$ and the dynamic adjustment amplitude $\beta$. The best results in each group are highlighted in \textbf{bold}.}
\label{tab:params}
\begin{tabular}{@{}cc|cccccccccc@{}}
\toprule
\multirow{2}{*}{$\lambda_{\mathrm{base}}$} & \multirow{2}{*}{$\beta$} & \multicolumn{2}{c}{\textbf{UAVTrack112}} & \multicolumn{2}{c}{\textbf{UAVDT}} & \multicolumn{2}{c}{\textbf{VisDrone2018}} & \multicolumn{2}{c}{\textbf{UAV123}} & \multicolumn{2}{c}{\textbf{Average}} \\
\cmidrule(l){3-4} \cmidrule(l){5-6} \cmidrule(l){7-8} \cmidrule(l){9-10} \cmidrule(l){11-12} 
& & Prec. & Succ. & Prec. & Succ. & Prec. & Succ. & Prec. & Succ. & Prec. & Succ. \\
\midrule
0.01 & 0.0 & \textbf{82.0} & \textbf{68.5} & 81.1 & 63.0 & 85.1 & 66.0 & 86.1 & 67.2 & 83.6 & 66.2 \\
0.10 & 0.0 & 81.8 & 68.2 & \textbf{81.5} & \textbf{63.4} & \textbf{85.6} & \textbf{67.6} & \textbf{86.5} & \textbf{67.3} & \textbf{83.9} & \textbf{66.6} \\
1.00 & 0.0 & 80.2 & 67.2 & 81.0 & 62.8 & 84.7 & 66.3 & 85.8 & 66.6 & 82.9 & 65.7  \\
\midrule
0.10 & 0.01 & 82.0 & 68.5 & 82.5 & 63.8 & 86.5 & 68.2 & 86.6 & \textbf{67.9} & 84.4 & 67.1 \\
0.10 & 0.05 & \textbf{83.0} & \textbf{69.7} & \textbf{84.0} & \textbf{65.3} 
& \textbf{89.1} & \textbf{69.4} & \textbf{86.7} & 67.6 & \textbf{85.7} & \textbf{68.0} \\
0.10 & 0.10 & 81.4 & 68.3 & 79.4 & 61.5 & 84.1 & 66.1 & 84.4 & 66.2 & 82.3 & 65.5\\
0.10 & 0.30 & 82.4 & 68.8 & 79.8 & 61.9 & 86.1 & 67.7 & 85.6 & 66.7 & 83.5 & 66.3\\
0.10 & 0.50 & 81.9 & 68.5 & 83.3 & 64.3 & 85.7 & 67.0 & 86.1 & 66.8 & 84.3 & 66.7\\
0.10 & 0.70 & 81.3 & 68.0 & 81.0 & 62.5 & 84.8 & 66.4 & 84.3 & 65.9 & 82.9 & 65.7\\
0.10 & 0.90 & 80.7 & 67.6 & 79.8 & 61.8 & 86.0 & 67.2 & 83.2 & 65.0 & 82.4 & 65.4\\
\bottomrule
\end{tabular}
\end{table*}
 
\subsection{Ablation Study}

\noindent\textbf{Effect of Mutual Information Maximization and Adaptive Weighting.}
To evaluate the effectiveness of Adaptive Mutual Information Maximization (AMIM), Table~\ref{tab:ablation} reports the results obtained by progressively introducing fixed-weight mutual information maximization (MIM) and adaptive dynamic weighting (ADW) into the baseline tracker. MIM consistently improves performance on all four benchmarks. Specifically, the average precision increases from 81.9\% to 83.9\%, while the average success rate increases from 65.2\% to 66.6\%, corresponding to gains of 2.0\% and 1.4\%, respectively. These improvements demonstrate that modeling the dependency between target-centered templates and their deep spiking representations benefits tracking performance.

Incorporating ADW further improves the results across all four benchmarks, yielding the highest average precision of 85.7\% and average success rate of 68.0\%. Compared with the fixed-weight MIM variant, ADW provides additional gains of 1.8\% in average precision and 1.4\% in average success rate. Overall, the complete AMIM improves the two metrics over the baseline by 3.8\% and 2.8\%, respectively. These results indicate that adapting the MI constraint to the relative localization difficulty of each training batch is more effective than using a fixed weighting scheme.

\noindent\textbf{Impact of AMIM Weighting Parameters.}
Table~\ref{tab:params} analyzes the sensitivity of AMIM to the base weight $\lambda_{\mathrm{base}}$ and the dynamic adjustment amplitude $\beta$, with the sensitivity factor $\eta$ fixed at 10 throughout the experiments.

We first set $\beta=0$ to isolate the effect of the fixed MI weight. Among the evaluated settings, $\lambda_{\mathrm{base}}=0.10$ achieves the best average performance, with a precision of 83.9\% and a success rate of 66.6\%. Reducing $\lambda_{\mathrm{base}}$ to 0.01 results in slightly lower average performance, indicating that an overly weak MI constraint provides limited regularization. In contrast, increasing it to 1.00 decreases the average precision and success rate to 82.9\% and 65.7\%, respectively, suggesting that excessive MI regularization may interfere with the primary localization objective.

We then fix $\lambda_{\mathrm{base}}=0.10$ and vary $\beta$ to evaluate the dynamic weighting strategy. The best overall performance is obtained at $\beta=0.05$, reaching an average precision of 85.7\% and a success rate of 68.0\%. Compared with the fixed-weight setting ($\beta=0$), this configuration improves precision and success rate by 1.8\% and 1.4\%, respectively. A smaller adjustment amplitude of $\beta=0.01$ provides more limited improvements, whereas larger values generally underperform the optimal setting. These results suggest that a moderate adjustment amplitude provides a better balance between the auxiliary MI constraint and the primary tracking objective.
\begin{figure*}[t]
    \centering
    \includegraphics[width=0.99 \textwidth]{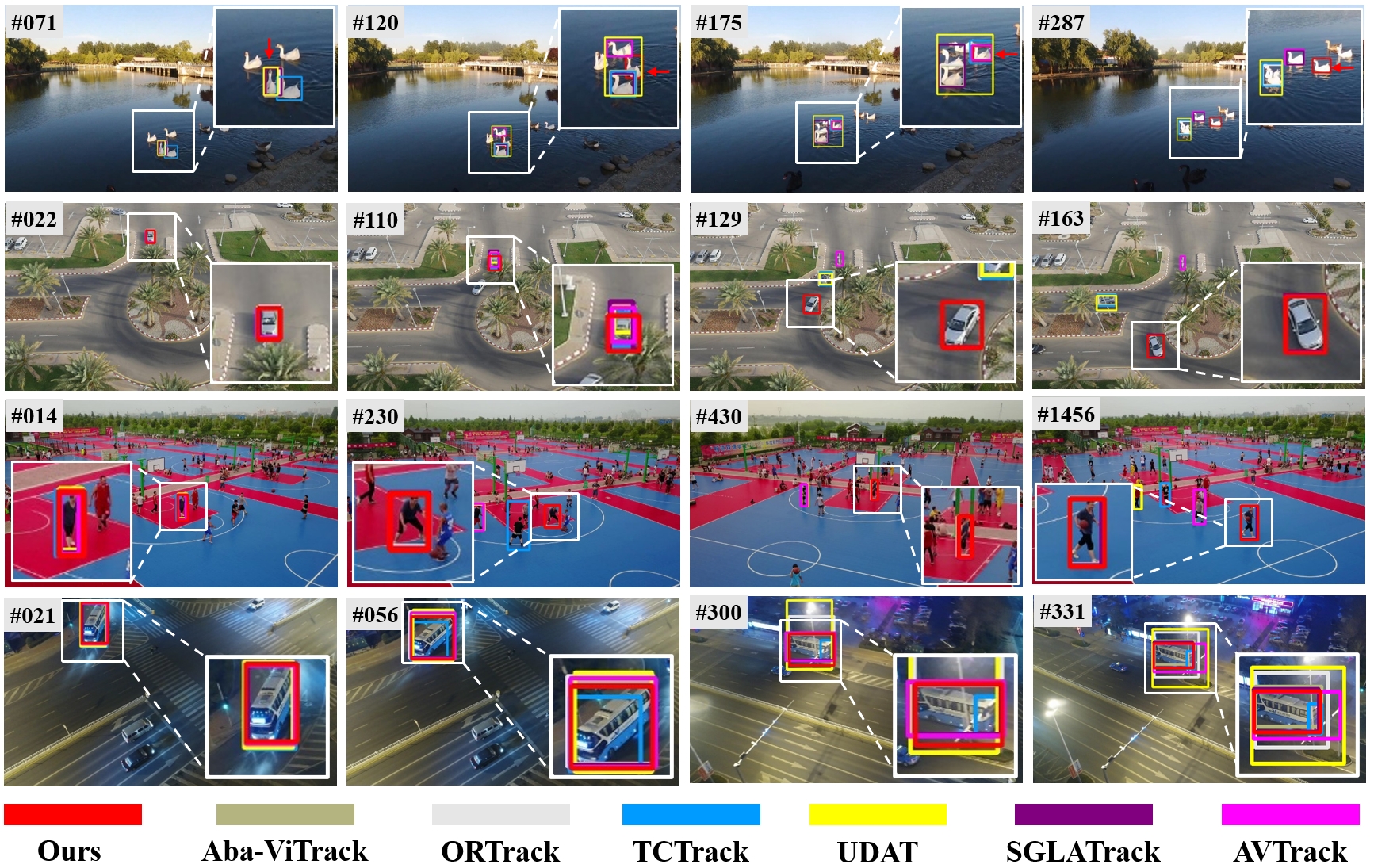}
    \caption{Qualitative comparison of STATrack with six state-of-the-art UAV trackers on four video sequences from UAVTrack112, UAV123, VisDrone2018, and UAVDT, namely \textit{swan}, \textit{Car7}, \textit{S1607}, and \textit{uav0000088\_00000\_s}.}
    \label{fig:track}
\end{figure*} 
\subsection{Qualitative Results}
Fig.~\ref{fig:feature} qualitatively compares the attention maps produced by the baseline and the proposed AMIM-enhanced model. The first four columns show representative frames from the \textit{uav0000074\_04320\_s} sequence in VisDrone2018, while the remaining four columns correspond to the \textit{uav4} sequence in UAV123. The first, second, and third rows present the search images, the response maps of the baseline, and those obtained with AMIM, respectively. In the \textit{uav0000074\_04320\_s} sequence, the target is surrounded by pedestrians and other visually similar objects. The baseline produces dispersed responses over both the target and nearby background regions, indicating considerable interference from surrounding distractors. In contrast, AMIM yields more compact and consistently target-centered responses across the displayed frames. A similar pattern can be observed in the uav4 sequence. Although the target drone is small and appears against varying backgrounds, the AMIM-enhanced model concentrates its responses more precisely on the target, whereas the baseline exhibits broader activation over background structures. These visualizations suggest that AMIM promotes more target-focused deep spiking representations while suppressing irrelevant background responses, supporting its effectiveness in improving the discrimination between the target and its surroundings.

Fig.~\ref{fig:track} presents a qualitative comparison of STATrack with six state-of-the-art UAV trackers on representative sequences from UAVTrack112, UAV123, VisDrone2018, and UAVDT. These sequences cover several challenging tracking conditions, including severe background clutter (\textit{swan}), interference from similar objects (\textit{swan} and \textit{Car7}), substantial scale variation (\textit{uav0000088\_00000\_s} and \textit{S1607}), and large appearance and pose changes. As shown in the figure, STATrack maintains accurate target localization under these challenging conditions, whereas several competing trackers suffer from target drift or tracking failure when confronted with background distractors, scale variation, or appearance changes. These qualitative results demonstrate the robustness of STATrack and suggest that its target-aware representations enable more stable tracking in complex UAV scenarios.

\section{Limitations}
\label{sec:limitations}

Although STATrack demonstrates a favorable balance between tracking performance and energy efficiency for RGB-based UAV visual tracking, several limitations remain. The reported energy consumption is mainly estimated based on theoretical MAC/AC operation costs and layer-wise spike firing rates. Although this method is widely adopted in SNN efficiency analysis, it cannot fully reflect the actual runtime energy consumption on specific hardware platforms. In particular, the runtime speed in this work is measured on conventional CPU/GPU platforms, which are primarily optimized for dense MAC operations rather than sparse spike-driven AC operations. Therefore, the hardware-level energy-efficiency advantage of STATrack still needs to be further validated on neuromorphic hardware or dedicated low-power accelerators. Future work will explore more adaptive and hardware-aware optimization strategies, as well as practical deployment on neuromorphic platforms, to further validate and enhance the applicability of SNN-based UAV trackers.

\section{Conclusion}

In this work, we presented STATrack, an RGB-only fully spiking neural network framework for energy-efficient UAV visual tracking. STATrack enables standard RGB images to be processed using a fully spiking tracking architecture without requiring event-camera input. To improve target-aware representation learning under limited-timestep spike-driven computation, we introduced Adaptive Mutual Information Maximization (AMIM), which models the statistical dependency between target-centered template inputs and their deep spiking representations. This constraint encourages the network to learn representations that better distinguish the target from its surrounding background. We further developed a sample-difficulty-aware weighting strategy that dynamically adjusts the strength of the MI constraint according to the relative localization difficulty of the current training batch. As AMIM is used only during training, it introduces no additional inference computation. Extensive experiments on four UAV tracking benchmarks showed that STATrack achieves the highest average success rate of 68.0\% among the compared methods, with a theoretical energy consumption of only 5.6 mJ. These results demonstrate the potential of fully spiking neural networks as an energy-efficient solution for RGB-based UAV visual tracking.

\section*{Acknowledgments}
This work was supported by the Guangxi Science and Technology Program (Grant Nos. GXYD2504850002 and XT2600870042).

\bibliographystyle{IEEEtran}
\bibliography{main.bib}

\newpage

\vfill

\end{document}